\pdfoutput=1

\documentclass[11pt]{article}

\usepackage[final]{acl}

\usepackage{times}
\usepackage{latexsym}
\usepackage{booktabs}

\usepackage[T1]{fontenc}

\usepackage[utf8]{inputenc}

\usepackage{microtype}

\usepackage{inconsolata}

\usepackage{graphicx}

\usepackage{hyperref}
\usepackage{url}
\usepackage{graphicx}
\usepackage{booktabs}
\usepackage{multirow, multicol}
\usepackage{diagbox}

\usepackage{ctable}
\usepackage{tabularx}
\usepackage{xspace}
\usepackage{amsmath}
\usepackage{float}
\usepackage{xcolor}
\usepackage{listings}
\usepackage{soul}
\usepackage{tablefootnote}
\usepackage{colortbl}

\usepackage{pifont}
\usepackage{amssymb}

\usepackage{enumitem}

\usepackage{listings}
\usepackage{listings-ext}
\usepackage{longtable}
\usepackage{placeins}
\usepackage{textcomp}

\usepackage{subcaption}
\usepackage{float}

\definecolor{mediumgreen}{RGB}{60, 179, 113}
\definecolor{darkgreen}{rgb}{0.0, 0.5, 0.0}

\lstdefinelanguage{Jinja2}{
  morekeywords={},
  sensitive=false,
  moredelim=[s][\color{blue}]{\{}{\}},
  moredelim=[s][\color{blue}]{\%}{\%},
  moredelim=[s][\color{mediumgreen}]{\{####}{####\}},
}

\lstdefinelanguage{ReAct}{
 basicstyle=\ttfamily\footnotesize,
  morekeywords={},
  sensitive=false,
  morecomment=[l][\color{darkgreen}\bfseries]{Thought:},
  emph={
    >>, search_wikidata, execute_sparql, get_wikidata_entry, stop
  },
  emphstyle={\color{blue}\bfseries},
  stringstyle=\color{red},
  morestring=[b]",
  morestring=[b]""",
    backgroundcolor=\color{lightgray!5},
    captionpos=b,
    tabsize=2,
    xleftmargin=2.0em,
    xrightmargin=2.0em,
    framexleftmargin=2.0em,
    framexrightmargin=2.0em,
    framextopmargin=1em,
    framexbottommargin=1em,
}

\definecolor{comment-red}{rgb}{0.8,0,0}
\definecolor{lightgray}{gray}{0.7}

\newcommand{\dataset}[0]{\textsc{Spinach}\xspace}
\newcommand{\system}[0]{\textsc{Spinach}\xspace}

\newcommand{\systemEmoji}{\includegraphics[height=1.1em,trim=5em 12em 7em 0]{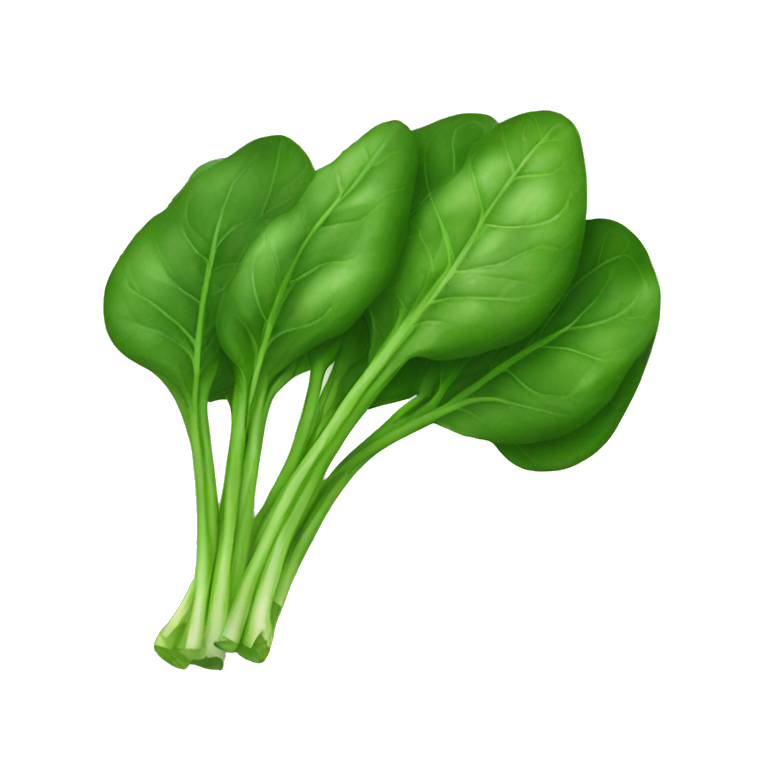}}
\newcommand{\systemWithEmoji}{\systemEmoji\xspace\system}

\title{\systemWithEmoji: \underline{SP}ARQL-Based \underline{I}nformation \underline{Na}vigation \\ for \underline{Ch}allenging Real-World Questions}

\author{
Shicheng Liu$^\dagger$\footnotemark[1] \quad 
Sina J. Semnani$^\dagger$\footnotemark[1] \quad
Harold Triedman$^\ddagger$$^\S$ \quad
Jialiang Xu$^\dagger$ \quad \\
\textbf{Isaac Dan Zhao$^\dagger$ \quad Monica S. Lam$^\dagger$}\\
$^\dagger$ Stanford University \quad $^\ddagger$ Cornell Tech \\
\fontsize{11}{12}\selectfont\texttt{\{shicheng, sinaj, xjl, ikezhao, lam\}@cs.stanford.edu}\\
\fontsize{11}{12}\selectfont\texttt{hjt36@cornell.edu}}

\begin{document}
\maketitle

\renewcommand{\thefootnote}{\fnsymbol{footnote}}
\footnotetext[1]{Equal contribution}

\renewcommand{\thefootnote}{\S}
\footnotetext[2]{Work conducted while at the Wikimedia Foundation}
\setcounter{footnote}{0}
\renewcommand{\thefootnote}{\arabic{footnote}}

\begin{abstract}
Large Language Models (LLMs) have led to significant improvements in the Knowledge Base Question Answering (KBQA) task. However, datasets used in KBQA studies do not capture the true complexity of KBQA tasks. They either have simple questions, use synthetically generated logical forms, or are based on small knowledge base (KB) schemas.

We introduce the \dataset dataset, an expert-annotated KBQA dataset collected from discussions on Wikidata's ``Request a Query'' forum with 320 decontextualized question-SPARQL pairs. The complexity of these in-the-wild queries calls for a KBQA system that can dynamically \emph{explore} large and often incomplete schemas and \emph{reason} about them, as it is infeasible to create a comprehensive training dataset. 

We also introduce an in-context learning KBQA agent, also called \system, that mimics how a human expert would write SPARQLs to handle challenging questions. \system achieves a new state of the art on the QALD-7, QALD-9 Plus and QALD-10 datasets by 31.0\%, 27.0\%, and 10.0\% in $F_1$, respectively, and coming within 1.6\% of the fine-tuned LLaMA SOTA model on WikiWebQuestions.
On our new \dataset dataset, the \system agent outperforms all baselines, including the best GPT-4-based KBQA agent, by at least 38.1\% in $F_1$.\footnote{Code and data available at \url{https://github.com/stanford-oval/spinach}. The \system agent is publicly available at \url{https://spinach.genie.stanford.edu/} and on Wikidata as \href{https://www.wikidata.org/wiki/User:SpinachBot}{SpinachBot}.}


  

\end{abstract}

\section{Introduction}

\begin{figure}[!ht]
    \centering
 \begin{subfigure}[t]{0.48\textwidth}
    \includegraphics[width=\textwidth]{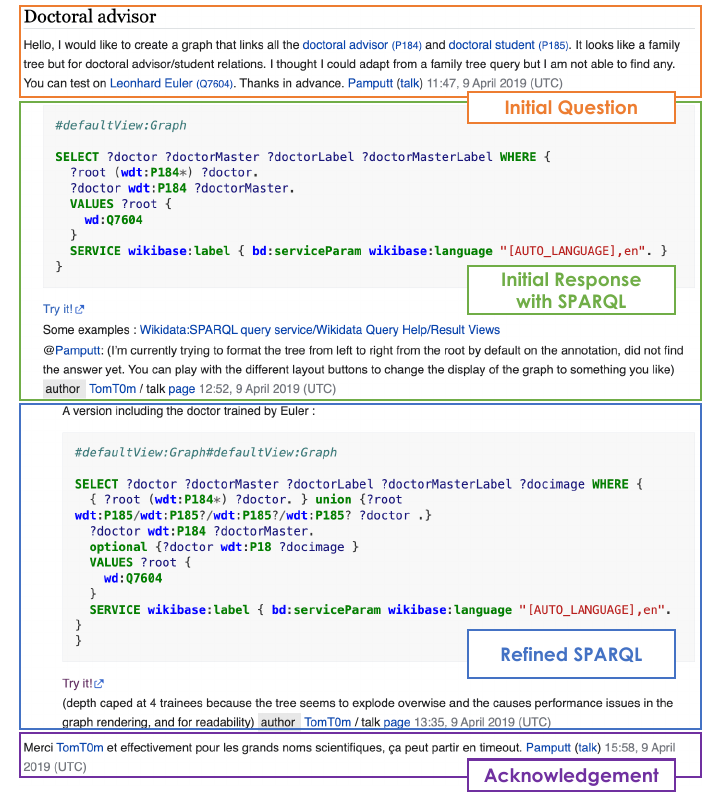}
    \caption{An example forum discussion}
    \label{fig:1a_forum_discussion}
\end{subfigure}

\vspace{1em} 

 \begin{subfigure}[t]{0.48\textwidth}
        \centering
        \resizebox{\textwidth}{!}{
        \begin{tabular}{l}
            \toprule
            \textit{\textbf{Question}}: Who are the doctoral advisors of Leonhard Euler,\\and their advisors, and so on? In addition, who are his doctoral student, \\grand-student, great-grand-student, and great-great-grand-students?\\ Each tuple in the result should contain both the student and the doctoral advisor. \\
            \midrule
            \textit{\textbf{SPARQL}}: \textbf{SELECT} ?doctor ?doctorMaster \textbf{WHERE} \{ \\
    ~~~~~\{ ?root (wdt:P184*) ?doctor. \} \\
    ~~~~~ \textbf{UNION} \\
    ~~~~~ \{ ?root (wdt:P185/(wdt:P185?)/(wdt:P185?)/(wdt:P185?)) ?doctor. \} \\
    ~~~~~ ?doctor wdt:P184 ?doctorMaster. \\
    ~~~~~ \textbf{VALUES} ?root \{ \\
    ~~~~~~~~~~ wd:Q7604 \\
    ~~~~~     \} \\
\} \\
            \bottomrule
        \end{tabular}
        }
        \caption{The corresponding decontextualized example included in the validation set of the \system dataset. The SPARQL is based on the ``Refined SPARQL'' proposed by the user \texttt{TomT0m}. Projection fields asking for labels and the optional image clause are removed in accordance with our methodology in Section~\ref{sec:dataset-inclusion}. } 
    \vspace{-1em}
    \label{fig:1b_forum_sparql}
    \end{subfigure}
\end{figure}

Wikidata~\citep{wikidata}, one of the largest publicly available knowledge bases, currently contains 15 billion facts and is estimated to grow at a rate of 1 billion triples per year~\citep{wikidata_sparql_query_service}. It is of significant value to many scientific communities, including Mathematics~\citep{scharpf2021mathematics}, Biology~\citep{Mitraka031971, Pfundner2015}, Education~\citep{EvensteinSigalov2023InvestigatingTP}, Linguistics~\citep{8308319,Yu2017MeronymyRE}, and the Social Sciences~\citep{wikidata-social-science-1, wikidata-social-science-2}, among many others~\citep{fardasarbas2019wikidata, turki2023novel}. Effective access to Wikidata data can be challenging. To address this, the Wikidata Request a Query forum\footnote{\url{https://www.wikidata.org/wiki/Wikidata:Request_a_query}} 
has been created so that users can ask questions, propose solutions, and participate in follow-up conversations. \subref{fig:1a_forum_discussion} shows one such conversation, in which a user wants to obtain the academic genealogy of Leonhard Euler. Note that in SPARQL, entities and properties are uniquely identified by QIDs and PIDs, respectively. Here P184 and P185 represent the ``doctoral advisor'' and ``doctoral student'', respectively, and Q7604 is Leonhard Euler. Readers are referred to \citet{wikidata_sparql_tutorial} for more information. 



Although numerous datasets have been proposed for Knowledge Base Question Answering (KBQA) task, they either contain only simple questions~\citep{yih-etal-2016-value, bordes2015largescale, qald7, qald-9-plus, qald10, xu-etal-2023-fine, RuBQ, RuBQ-2} or synthetically generated complex logical forms~\citep{bao-etal-2016-constraint, grailQA, talmor-berant-2018-web, keysers2020measuring, cao-etal-2022-kqa, lc-quald-2.0}. Datasets with synthetically generated logical forms often lead to an overestimate of the performance of KBQA systems;  performing well on them does not translate to real-world queries~\cite{oren-etal-2021-finding, campagna-etal-2022-shot}. Thus, the community needs a high-quality dataset with organic, real-world queries that capture the true complexity of KBQA tasks.

In this paper, we propose using the Wikidata Request a Query forum to build and evaluate next-generation KBQA systems. \textbf{We introduce the \systemWithEmoji dataset, a new, expert-annotated KBQA dataset featuring decontextualized question-SPARQL pairs derived from complex discussions on a real-world forum.} \subref{fig:1b_forum_sparql} is the example obtained from the discussion in \subref{fig:1a_forum_discussion}. 


As we show in this paper, current KBQA approaches~\citep{xu-etal-2023-fine, sun2024thinkongraph} cannot handle the complexity of these real-world queries. The state-of-the-art (SOTA) approach, ToG~\cite{sun2024thinkongraph}, which integrates LLMs with KG reasoning, scores only 1.8 EM and 7.2 F1 on this data set. 
Therefore, {\bf we propose a new LLM-augmented KBQA approach, \systemWithEmoji}: \textbf{SP}ARQL-Based \textbf{I}nformation \textbf{Na}vigation for \textbf{Ch}allenging Real-World Questions, which is designed with the primary goal of mimicking how an expert would tackle the task. We show that \system establishes a new state of the art on popular datasets such as QALD-7, QALD-9, and QALD-10 and comes within 1.6\% of the fine-tuned SOTA on WikiWebQuestions. 
On the \dataset dataset, our agent outperforms
all baselines, including the best GPT-4-based
KBQA agent, by at least 38.1\% in F1.


\section{Related Work}

\subsection{KBQA Benchmarks}

\begin{table*}[htbp]
\small
    \centering
    \resizebox{\textwidth}{!}
{
    
    \begin{tabular}{l|ccccccc}
        \toprule
     &  Avg. Clauses & Avg. Projs & Avg. Rels & Avg. Subjs & Avg. Preds & Avg. Objs & Avg. Lits \\ 
        \toprule
    \multicolumn{8}{c}{\textbf{Natural questions w/ annotated logical forms}} \\
    \midrule
  WikiWebQuestion~\citep{xu-etal-2023-fine} & 2.63 & 1 & 1.53 & 1.25 & 1.52 & 1.53 & 0.04\\
  QALD-9 Plus~\citep{qald-9-plus} & 3.14 & 1  & 1.77 & 1.26 & 1.70 & 1.78 & 0.05 \\
  QALD-10~\citep{qald10} &2.38  & 1& 1.27  & 1.19 & 1.17 & 1.32 & 0.05 \\
  RuBQ~\citep{RuBQ} & 2.17 & 1  & 1.12 & 1.03 & 1.11 & 1.07 & 0.01 \\
  SimpleQuestionsWikidata~\citep{simplequestionswikidata} & 2.00  & 1  & 1.00 & 1.00 &1.00 & 1.00 & 0.00 \\
  \midrule
  \multicolumn{8}{c}{\textbf{Synthetic logical forms w/ synthetic or paraphrased questions}} \\
  \midrule
  CWQ~\citep{talmor-berant-2018-web} & 5.19 & 1  & 2.80 & 1.87 & 2.62 & 3.38 & 0.11 \\
  GrailQA~\citep{grailQA} & 7.10 & 1  & 3.02 & 1.97 & 2.43 & 3.90 & 0.08 \\
  KQA Pro~\citep{cao-etal-2022-kqa} & 6.34 & 1 & 5.01 & 2.77 & 3.94 & 2.43 & 2.37 \\
  MCWQ~\citep{cui-etal-2022-compositional} & 6.34 & 1  & 5.09 & 2.67 & 3.53 & 3.37 & 0.00 \\
  LC-QuAD-2~\cite{lc-quald-2.0} &3.65 & 1 & 2.07  & 1.51  &2.05  &2.07  & 0.22 \\
  \midrule
    \multicolumn{8}{c}{\textbf{Natural logical form w/ annotated questions}} \\
\midrule
  \systemWithEmoji \textbf{(Ours)} & 8.89 & 2.50 & 4.03 & 1.76 & 3.55 & 4.53 & 0.46 \\
        \bottomrule
    \end{tabular}

}
    \caption{Quantitative comparison of KBQA datasets on average number of Clauses, Projections, Relations, Subjects, Predicates, Objects, and Literals. For datasets originally based on Freebase, we calculate their corresponding datasets in Wikidata, if available. Refer to Appendix \ref{appendix:dataset-metrics-definition} for definitions of each metric.}
    \label{tab:dataset-analysis}
\end{table*}

Over the past decade, numerous Knowledge Base Question Answering (KBQA) benchmarks with logical forms have been introduced and can be classified into two categories:

\textbf{Datasets with natural questions originally collected through search engines or crowd-sourcing.} This includes the popular dataset WebQuestionSP~\citep{yih-etal-2016-value}, along with the QALD datasets~\cite{qald7, qald-9, qald10, qald-9-plus}, RuBQ~\citep{RuBQ}, and SimpleQuestions~\citep{bordes2015largescale}, among others;

\textbf{Datasets with synthetically generated logical forms and questions}, where some datasets paraphrase synthetically generated questions via crowdsourcing. This category includes ComplexWebQuestions~\citep{talmor-berant-2018-web}, GrailQA~\citep{grailQA}, KQA Pro~\citep{cao-etal-2022-kqa}, CFQ~\citep{keysers2020measuring}, and LC-QuAD-2~\cite{lc-quald-2.0}, among others.

Some of the aforementioned datasets were originally based on Freebase~\citep{freebase} or DBPedia~\cite{DBPedia}. Recently, however, the KBQA community has shifted toward using Wikidata as the underlying knowledge base for KBQA datasets, due to its larger size and continuous updates from community contributions.\footnote{For more on why Wikidata is better suited for KBQA benchmarks, refer to Section 1.1 of \citet{qald10}.} Several datasets have been converted to use Wikidata: WebQuestionSP has been converted to WikiWebQuestions~\citep{xu-etal-2023-fine}; SimpleQuestions has a Wikidata version~\citep{simplequestionswikidata}; and CFQ has been converted to MCWQ~\citep{cui-etal-2022-compositional}.

In Table \ref{tab:dataset-analysis}, we present quantitative statistics on these datasets. Existing datasets with natural questions typically involve relatively simple queries (e.g., 2.63 clauses per query for WikiWebQuestions). In contrast, synthetically generated datasets generally contain more complex queries by design. However, due to the limited natural language variety between training and evaluation data, models can achieve artificially high accuracy. For instance, a simple semantic parser based on the BART~\cite{lewis-etal-2020-bart} model can achieve an accuracy of over 90\% on KQA-Pro even without doing entity linking~\citep{cao-etal-2022-kqa}. Furthermore, systems that excel at synthetic datasets do not necessarily perform well in natural, complex tasks~\citep{oren-etal-2021-finding, campagna-etal-2022-shot}. Additional details on the number of properties and unique query patterns of existing datasets and \dataset are presented in Appendix \ref{appendix:additional-comparison-with-synthetic-data}. This shows the need for a high-quality dataset with both natural questions and natural complex logical forms in the KBQA community.


\subsection{KBQA Approaches}
\label{sec:kbqa-approaches}
Current KBQA systems can be classified into three categories: (1) subgraph retrieval using vector embeddings~\citep{sun-etal-2018-open, sun-etal-2019-pullnet, sen-etal-2021-expanding, verga-etal-2021-adaptable, mavromatis-karypis-2022-rearev}, (2) semantic parsing~\citep{yih-etal-2015-semantic, yih-etal-2016-value, luo-etal-2018-knowledge, lan-jiang-2020-query, das-etal-2021-case, ye-etal-2022-rng, cao-etal-2022-program, gu-su-2022-arcaneqa, xu-etal-2023-fine}, and more recently, (3) LLM-based graph exploration~\citep{sun2024thinkongraph, xiong2024interactivekbqa}. Some works use a combination of these techniques~\citep{yu2023decaf, luo2024chatkbqa, luo2024reasoning}.

Due to the sheer size of Wikidata, embedding its entire graph is prohibitively expensive. Subgraph retrieval approaches therefore limit the problem to a small fixed subset of Wikidata. \citet{xiong2024interactivekbqa}, for instance, create and search through embeddings of the graph patterns (triples) for relevant patterns and conduct experiments on a tiny subgraph covering only 0.01\% of Wikidata (i.e. containing 17K QIDs instead of the full 111,568K). This limitation renders them inapplicable to real-world scenarios as it leaves much of the knowledge graph unaccessible to users. Among semantic parsing systems, \citet{xu-etal-2023-fine} fine-tune LLaMA~\cite{touvron2023llama} with a modified SPARQL syntax and achieve state-of-the-art results on 2 KBQA benchmarks. Among LLM-based approaches, \citet{sun2024thinkongraph} instruct an LLM to dynamically explore the graph to fetch answers, achieving SOTA on 6 KBQA datasets. In Section~\ref{sec:experiments}, we evaluate these two systems on our new \dataset dataset.

\section{The \dataset Dataset}

The archives of the Wikidata Request a Query from July 2016 to the present date are available\footnote{\url{https://www.wikidata.org/wiki/Wikidata:Request_a_query/Archive}}.
These conversations are real and organic, reflecting the kind of queries that practitioners are interested in. Additional details on the forum and analysis of the conversations can be found in Appendix \ref{appendix:additional-detail-request-a-query}.

Out of the discussions up to May 2024, 2780 discussions include at least one valid SPARQL query. We keep conversations whose last-mentioned SPARQL query returns non-empty results under 10 MB. Of the 2171 discussions left, we randomly sample 920 conversations spanning many domains for consideration.
These conversations typically do not specify the exact natural language corresponding to the SPARQLs. Thus, manual processing is required to convert them to a KBQA dataset. Three Wikidata experts among the authors of this paper manually inspected these conversations to produce a dataset with 155 examples in the validation and 165 examples in the test set, as described below.

Additional discussions on the size and statistical power of \dataset dataset can be found in Appendix \ref{appendix:size-discussion}. More information on the process of converting source conversations and SPARQLs into a KBQA dataset can be found in Appendix \ref{appendix:annotation-details}. 

\subsection{Choosing and Trimming the Queries}
\label{sec:dataset-inclusion}
We design the dataset to focus on end-users rather than Wikipedia and Wikidata contributors interested in obscure optimizations or formatting. As such, we manually remove the following SPARQL clauses:

\textit{Wikimedia presentation queries}: We filter out clauses for analyzing or improving Wikimedia projects themselves, such as asking which Wikipedia articles exist in one language but not another.

\textit{Questions on complex SPARQL code}. We exclude conversations where users ask for help with debugging overly complicated SPARQL queries, when their meaning is already difficult to accurately convey in natural language.

\textit{Queries obscured by optimizations}: Because Wikidata restricts query runtime to 60 seconds~\citep{wikidata-timeout}, users may include clauses to optimize the performance, which are often not directly relevant to answering the question. We remove these clauses.

\textit{Formatting clauses}: We remove clauses that only format the results.

Appendix \ref{appendix:examples-of-queries-modified} and \ref{appendix:examples-of-queries-excluded} show specific examples of these modifications and exclusions.

\subsection{Annotating Natural Questions}

For each conversation, the experts annotate a self-contained, decontextualized natural language question that \emph{accurately} captures the meaning of the user-written SPARQL. We follow these steps:

\textit{Disambiguate entities and properties}: Entities and properties are important parts of a query. For ambiguous entities, the verbalization includes additional information to resolve the ambiguity. As for the properties, the verbalization should distinguish between similar properties. For example, instead of asking ``where a movie takes place'', we distinguish between the ``narrative location'' and the ``filming location''; instead of asking ``where a person comes from'', we distinguish between the ``country of citizenship'' vs. ``country of birth''. 


\textit{Natural verbalizations}: Whenever possible, to the extent that no ambiguities are introduced, the question should refrain from directly using entity and property names, instead using a more natural way to express the meaning. For instance, instead of asking ``what is the point of time of the goal?'', a more natural question with the same level of accuracy like ``when does the goal take place?'' should be used.

\textit{Accurately capturing optional clauses and projections}: In SPARQL, \texttt{OPTIONAL} clauses are used to include additional data that may or may not exist for queried items without excluding results lacking this optional data.\footnote{\url{https://www.wikidata.org/wiki/Wikidata:SPARQL\_tutorial\#OPTIONAL}} The verbalization should specify which fields are optional with clauses such as ``if available''. Similarly, the verbalization should accurately capture what is requested in the projections, using phrases like: ``For each result, return their name and location'' if necessary.
 
As illustrated in Table \ref{tab:dataset-analysis}, the \dataset dataset is the first dataset with both natural and complex logical forms, which represents the real-world KBQA needs of Wikidata users. Relative to previous datasets, there are more unique properties and unique logical forms per query. Additional details on the popularity of the query topics in the \dataset dataset can be found in Appendix \ref{appendix:analysis-popularity-of-queries}.


\section{The \system Agent}

The \system agent is an LLM-augmented knowledge graph exploration agent, where an LLM is instructed to explore the knowledge graph until an appropriate answer is found or another stop condition is met. 
However, unlike prior work, we design \system with the primary goal of mimicking a human expert writing a SPARQL query. An expert starts by writing simple queries and looking up Wikidata entity or property pages when needed, all to understand the structure of the knowledge graph and what connections exist. This is especially important for Wikidata due to its anomalous structure~\cite{quality-of-wikidata-2022}.
An expert then might add new SPARQL clauses to build towards the final SPARQL, checking their work along the way by executing intermediate queries and eyeballing the results. Potentially, they might go back to the drawing board and double-check their assumptions if a query fails.

Following this design principle, \system improves upon the following aspects of previous agent-based KBQA systems:

First, \system uses the full expressiveness of SPARQL for exploration. This contrasts with \citet{sun2024thinkongraph, xiong2024interactivekbqa, sun2024oda}, which explore the knowledge graph one edge at a time. 
That is, \system weaves together the exploration and the writing of SPARQL. It is instructed to \emph{try many SPARQLs and fail early}. It learns about the structure of the knowledge graph from the results of its queries (or lack thereof), or realizes its incorrect assumptions, and decides to revisit them. This is a key difference from \citet{sun2024oda, sun2024thinkongraph}, where the answer is generated only when exploration is done.

Second, during its exploration, \system does not keep track of a subgraph as its state. Instead, we define the state to be the full history of actions and their results so far. Limiting the state to a subgraph~\cite{sun2024oda, sun2024thinkongraph} means that the agent would categorically not support questions whose answer contains a large number of entities (e.g. ``Which actors graduated from Harvard?'') or computation (e.g. ``What is the tallest mountain?'').

Third, while most KBQA agents start their exploration from entities they detect from the question, \system does not assume access to entities. This, for example, gives the agent the flexibility to start by searching for relevant relations if that is a more suitable way to explore.

\begin{figure}[!ht]
\includegraphics[width=0.48\textwidth]{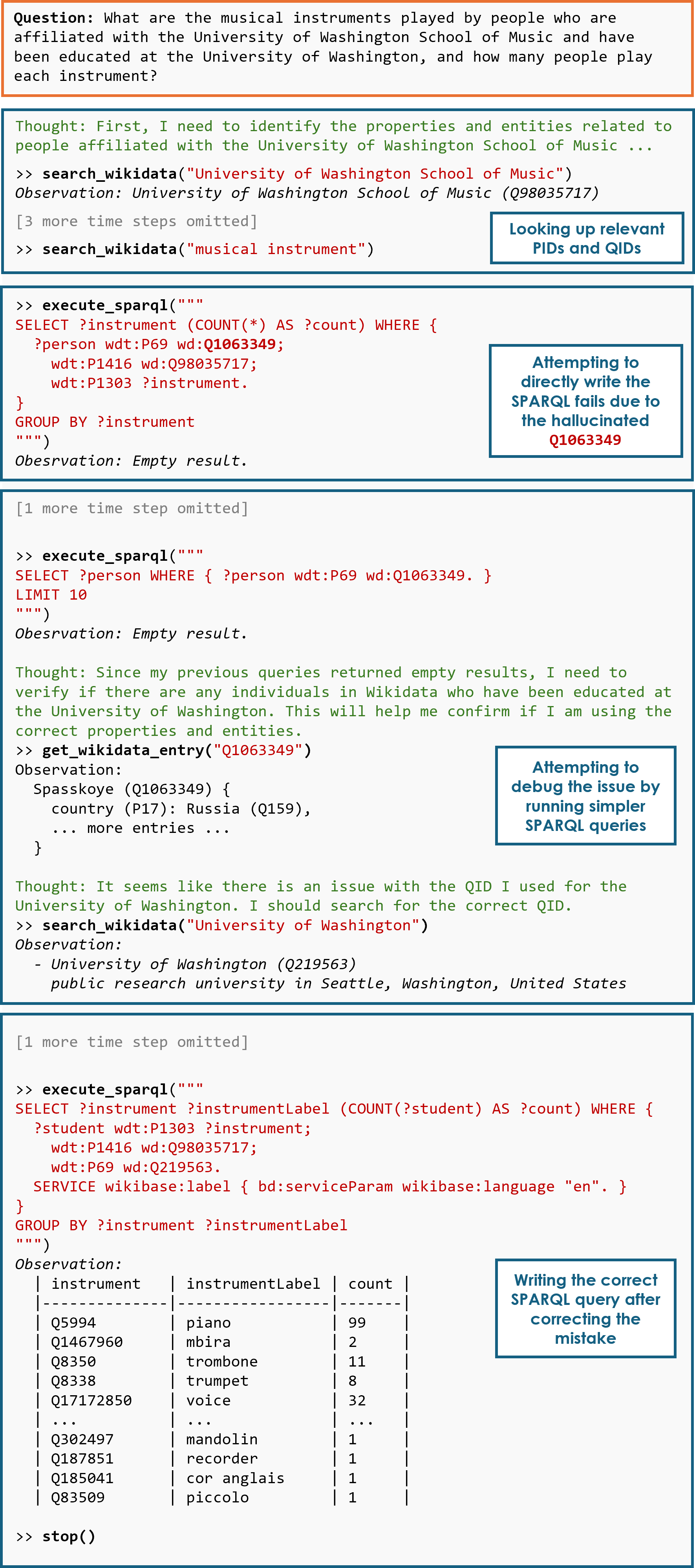}

\caption{The sequence of 13 actions that the \system agent takes to answer a sample question from the \dataset validation set. Here, the agent goes through several distinct phases, only with the high-level instruction in Section~\ref{sec:policy}.
Note that every step includes a thought, action and observation, but some are omitted here for brevity. Full version available in Listing~\ref{fig:full_example}.}
\label{fig:example}
\vspace{-1.5em}
\end{figure}

Concretely, \system agent runs for $N$ time steps. At time step $n$, it takes the current \emph{state} as input and outputs a \emph{thought}~\cite{yao2023react} $t_n$ and one of the possible actions $a_n$ from the set of all possible actions. The action is then executed outside of the agent, and the results are returned to the agent as observation $o_n$.
In the rest of this section, we go into the details of the \system agent. Figure~\ref{fig:example} shows an example of how \system answers a question.

\subsection{Exploration State}
The exploration state at time step $n$ is $\mathcal{S}_n=\{t_1, a_1, o_1, ..., t_{n-1}, a_{n-1}, o_{n-1}\}$, that is, the list of all thoughts, actions, and observations up to and including the previous time step. $\mathcal{S}_1$ is set to $\emptyset$.

Complex KBQA tasks require high reasoning ability. One promising approach to improve the reasoning capabilities of LLMs is ReAct prompting~\cite{yao2023react}, in which the LLM is instructed to output its reasoning trace in natural language first, before selecting an action. We require the agent to output a \emph{thought} at the beginning of each time step.

At time step $n$, the \system agent takes $a_n$, one of the following five actions, and receives the result of the execution as observation $o_n$. These actions are selected for their similarity to what a human expert can do.

\noindent\textbf{\texttt{search\_wikidata(string)}}
is equivalent to using the text search box at \url{wikidata.org}. This action searches Wikidata for items (entities or properties) that match a given string using the Wikidata API\footnote{\url{https://www.wikidata.org/w/api.php}}
\texttt{wbsearchentities}. This API finds matches using human-readable labels and aliases. The main use for this action is to find relevant QIDs and PIDs.

\noindent
\textbf{\texttt{get\_wikidata\_entry(QID)}}
is equivalent to visiting the Wikidata page for an entity, e.g. \url{https://www.wikidata.org/wiki/Q219563}.
This action retrieves all outgoing edges, i.e. linked entities, properties, and qualifiers of a specified Wikidata entity using its QID.

\noindent\textbf{\texttt{get\_property\_examples(PID)}} provides a few examples demonstrating the use of the specified property in Wikidata. The main use of this action is to understand how a property is used in Wikidata, if it is not clear from its label alone.

\noindent\textbf{\texttt{execute\_sparql(SPARQL)}}
is equivalent to using the Wikidata Query Service at \url{https://query.wikidata.org/}. It runs a SPARQL query on Wikidata and returns its results.
Executing this action gives the agent one of the following four types of feedback signal:
\begin{enumerate}[itemsep=-0.5em, topsep=0.5em]
    \item \textit{The query returns non-empty results.} We have observed that LLM is capable of reasoning over these results to choose the next action.
    \item \textit{The query returns empty results.} We have observed that the agent often attempts to debug the query clause-by-clause or attempts to take alternative actions.
    \item \textit{The query has a syntax error.} We show the specific syntax error message that the Wikidata Query API returns to the agent. We have observed that the agent can use the syntax error message to correct its previous syntactically incorrect SPARQL.
    \item \textit{The query execution times out.} The Wikidata Query API caps the execution at 60 seconds. When a timeout occurs, we have observed that the agent attempts to simplify the SPARQL query in some way to avoid timeouts
\end{enumerate}

\noindent\textbf{\texttt{stop()}}
marks the last executed SPARQL query as the final answer and ends the process.

Importantly, the results of the execution of each action are put in a human-readable format to make it easier for the LLM to process. To limit the amount of information that the agent has to process, we limit the output of search results to at most 8 entities and 4 properties, and limit large results of SPARQL queries to the first and last 5 rows. The outputs of \texttt{get\_wikidata\_entry()} are pruned using a separate few-shot prompt (Prompt ~\ref{prompt:prune}) to only keep entries that are relevant to the question.

\subsection{Policy}
\label{sec:policy}
The policy of \system is implemented using a zero-shot prompt that only includes high-level instructions such as
``start by constructing very simple queries and gradually build towards the complete query''
and ``confirm all your assumptions about the structure of Wikidata before proceeding'' (Prompt \ref{prompt:controller}). The decision of selecting the action at each time step is left to the LLM. This decision is made after experimenting with more restrictive policies, but we empirically observe that accounting for all various edge cases might confuse the agent and hurt the performance.

In each round of exploration, the agent constructs the exploration state and uses the policy prompt to ask LLM to output a thought and an action. It then executes the action and adds the observation to the state. In practice, the LLM can occasionally (1) fall into a cycle of repeating the same action and argument over and over again, or (2) deviate from its instructions and call \textbf{\texttt{stop}} even though the last SPARQL execution did not return any results. To remedy this, the agent checks for these potential problems and resets the exploration state to the one before the repetition or the deviation, and continues  from there.

Note that the policy LLM is run using sampling with high temperature of 1.0 to encourage exploration, so rerunning from a previous state does not repeat the exact same sequence of actions. The agent loop continues until either (1) the \texttt{stop()} action is selected, or (2) 15 actions is taken after taking ``rollbacks'' into account, or (3) a total of 30 actions.\footnote{The agent has a budget of 15 actions to take, and an extra 15 actions to spend on these ``rollbacks'', to a total of 30 actions. For instance, the agent can take 9 actions, get rolled back by 4 actions, then take another 10 actions and stop. This means the final solution has $9-4+10=15 \le 15$  actions, but the total number of actions is $9+4+10=23\le 30$}. 

In the example in Figure~\ref{fig:example}, \system goes through the following four phases, just by following the high-level instruction we provided:
it (1) looks up relevant PIDs and QIDs, (2) attempts to write the SPARQL in one go, which fails due to a hallucinated QID that the agent did not previously check, (3) starts the debugging process by executing simpler SPARQL queries, and double-checking its assumptions. It realizes its mistake after looking at the result of a \texttt{get\_wikidata\_entry}, and finally (4) moves on to write the correct SPARQL query.

\paragraph{Discussion} In our initial experiments with imposing low-level control over the selected actions, we found that it does not improve the accuracy of the agent. There are many possible states that the agent can encounter, and programming the best action in each one is challenging, especially in our zero-shot setting and without methods that can directly learn from large amounts of policy data.

Therefore, we leave most of the action selection to the LLM, and only impose have high-level control via the ``policy prompt'' (Prompt \ref{prompt:controller}) with instructions like ``Confirm all your assumptions about the structure of Wikidata before proceeding.'', which encourages \textbf{\texttt{search\_wikidata}} and \textbf{\texttt{get\_wikidata\_entry}} in the beginning, before executing any SPARQLs via \textbf{\texttt{execute\_sparql}}. This way, we leverage the underlying LLM’s common sense and world knowledge to reason over its observations. For example, we saw empirically that the LLM is capable of determining if the result of a SPARQL query appears to be implausible (e.g. returning a long list of entities when the question is about “the oldest”).

\section{Experiments}
\label{sec:experiments}
\begin{table*}[ht]
\small
    \centering
    \resizebox{\textwidth}{!}{
    \begin{tabular}{l|cc|cc|cc|cc|cc|cc}
        \toprule
    & \multicolumn{2}{c|}{QALD-7 (Task 4)} & \multicolumn{2}{c|}{QALD-9 Plus (en)} & \multicolumn{4}{c|}{QALD-10 (en)} & \multicolumn{4}{c}{WikiWebQuestions} \\
    & \multicolumn{2}{c|}{Test}  & \multicolumn{2}{c|}{Test} & \multicolumn{2}{c|}{Full Test Set} & \multicolumn{2}{c|}{Subset in ToG} &\multicolumn{2}{c|}{Dev} &\multicolumn{2}{c}{Test} \\
    & EM & F1 & EM & F1 & EM & F1 & EM & F1 & EM & F1 & EM & F1 \\
    \midrule
STAGG~\cite{yih-etal-2016-value} & - & 19.0 & - & - & - & - & - & - & - & - & - & - \\
GGNN~\cite{sorokin-gurevych-2018-modeling} & - & 21.3 & - & - & - & - & - & - & - & - & - & - \\
LingTeQA~\cite{qald-7-to} & - & 34.0 & - & - & - & - & - & - & - & - & - & - \\
\citet{qald10-baseline-baramiia2022ranking} & - & - & - & -& - & 42.8 & - & - & - & - & - & - \\
\citet{qald10-baseline-shivashankar2022graph} & - & - & - & -& - & 49.1 & - & - & - & - & - & - \\
QAnswer~\citep{qald10-baseline-diefenbach2017wdaqua} & - & 40.0& - & 44.6 & - & 57.8 & - & - & - & - & - & - \\

SPARQL-QA~\citep{qald10-baseline-borroto2022sparql} & - & - & - & -& - & 59.5 & - & - & - & - & - & - \\

\citet{liu2024enhancing} & - & - & - & -& 56.5 & - & - & - & - & - & - & - \\
0-shot ToG (GPT-4)~\citep{sun2024thinkongraph} & - & - & - & -& - & - & 54.7 & - & - & - & - & - \\
Fine-tuned WikiSP~\citep{xu-etal-2023-fine} & 38.0 & 43.6 & - & -& - & - & - & - & \textbf{75.6} & \textbf{76.9} & \textbf{65.5} & \textbf{71.9} \\
    \midrule
0-shot \dataset agent (GPT-4o)~\textbf{(Ours)} & \textbf{62.2} & \textbf{74.6} & \textbf{58.3} & \textbf{71.6} & \textbf{63.1} & \textbf{69.5} & \textbf{64.7} & \textbf{72.4} & 61.2 & 72.3 & 59.9 & 70.3 \\
        \bottomrule
    \end{tabular}
    }
    \caption{Performance of the \system agent and prior works on 4 prior datasets. \citet{sun2024thinkongraph} only evaluated on the subset of non-boolean questions of QALD-10 test set, which we denote as ``Subset in ToG''. \system achieves the new SOTA on QALD-7, QALD-9 Plus, and QALD-10. On WikiWebQuestions, it comes within 1.6\% F1 to the SOTA WikiSP fine-tuned on the dataset. } 
    \label{tab:prior-results}
\end{table*}

\begin{table*}[ht]
\small
    \centering
    \begin{tabular}{l|rr|rr}
        \toprule
    &  \multicolumn{2}{c|}{Dev} & \multicolumn{2}{c}{Test}\\
    & EM & F1 & EM & F1 \\
    \midrule
Direct GPT-4o Question Answering &0.0 &3.9 &0.0 &4.0 \\
GPT-4o Generating SPARQL & 1.3 & 5.4 & 0.6 & 3.9 \\
Fine-tuned WikiSP~\citep{xu-etal-2023-fine} & 1.3& 3.5& 1.2 &7.1 \\
0-shot ToG (GPT-4) ~\citep{sun2024thinkongraph} & 3.9 & 9.8 & 1.8 & 7.2 \\
\midrule
0-shot \dataset agent (GPT-4o)~\textbf{(Ours)} & \textbf{21.4} & \textbf{46.4} & \textbf{16.4} & \textbf{45.3} \\
        \bottomrule
    \end{tabular}
    \caption{Evaluation of the \system agent and prior works on the \system dataset.} 
\label{tab:prior-works-on-spinach}
\end{table*}

\subsection{Evaluation Metrics}

Prior works mostly use two metrics: Exact Match (EM) and $F_1$~\citep{yih-etal-2016-value, qald10}. 
As shown in Table \ref{tab:dataset-analysis}, the \dataset dataset contains 2.5 projections on average for each query, whereas prior datasets all only contain only one field in projection. This means that naively applying EM and $F_1$ is not possible; therefore, we propose a generalization of these metrics.

From the user's perspective, answers containing more than the minimum information which help clarify or enrich the answer are welcome. For the query ``what is the county with most people in South Dakota?'', a LLM-based system can choose to return the top county along with its population, even though the gold answer contains only the county. In other words, answers that contain additional projection columns from the gold should not be penalized.
To reflect this, we introduce a \textbf{row-major} generalization of EM and F1 to handle matrix-wise comparisons in real-life KBQA tasks, where each row is andled such that extra columns are not penalized. 

Consider the general case where the answer of a query consists of $m$ projections of $n$ results. Let 
\[
\scalebox{0.87}{
\(
\mathbf{y} = \begin{pmatrix}
y_{11} & \cdots & y_{1m} \\
\vdots  & \ddots & \vdots \\
y_{n1}  & \cdots & y_{nm} \\
\end{pmatrix}
\hspace{0.5cm}
\mathbf{y'} = \begin{pmatrix}
y'_{11} & \cdots & y'_{1m'} \\
\vdots  & \ddots & \vdots \\
y'_{n'1}  & \cdots & y'_{n'm'} \\
\end{pmatrix}
\)
}
\]
be the gold and predicted result, respectively. 

We first define the recall between a row in gold and a row in predicted such that additional columns in the predicted result are not penalized.
Let $y_i$ and $y'_j$ be the $i$ and $j$ rows in the gold and predicted results, respectively. 
$$\text{recall}(y_i, y'_j) = \frac{|y_i \cap y'_j|}{|y_i|}$$

To calculate the recall between the full gold and the predicted answer, we first find assignment~\citep{scipy_linear_sum_assignment} $A(\mathbf{y}, \mathbf{y'})
=\{(i_1,j_1),...,(i_r,j_r)\}$ where row $y_{i_k}$ is matched with $y'_{j_k}$ such that the sum of the recall between the matched rows is maximized.
(Matching rows with 0 recall is disallowed.)  

For the calculation of $F_1$, 
given $\mathbf{y}$ with $n$ rows, $\mathbf{y'}$ with $n'$ rows, and $A(\mathbf{y},\mathbf{y'})=\{(i_1,j_1),...,(i_r,j_r)\}$, the number of true positives, false negatives, and false positives are, respectively:
\[
\textit{tp} = \sum_{(i,j)\in A(\mathbf{y},\mathbf{y'})} \text{recall}(y_i, y'_{j})
\]
\[
\textit{fn} = n - r + \sum_{(i,j)\in A(\mathbf{y},\mathbf{y'})} 1-\text{recall}(y_i, y'_{j}) 
\]
\[
\textit{fp} = n' - r
\]

The row-major $F_1$ between $\mathbf{y}$ and $\mathbf{y'}$ is:
\[
\frac{2\textit{tp}}{2\textit{tp} + \textit{fp} + \textit{fn}}
\]

EM is defined to be $1$ if the row-major $F_1$ score is $1$ and $0$ otherwise.
Note that with this definition, EM and $F_1$ are exactly equal to the tradition al definition of EM and $F_1$ when there is only one projection.

\subsection{\dataset Agent on Prior Datasets}

We evaluate our approach on four previous Wikidata datasets.
We experiment with three QALD challenges that have annotated Wikidata SPARQLs: QALD-7 (task 4 for Wikidata)~\citep{qald7}, QALD-9 Plus~\cite{qald-9-plus}, and the English subset of QALD-10~\citep{qald10}. In addition, we also evaluate with WikiWebQuestions~\citep{xu-etal-2023-fine}, the Wikidata version of the popular Web\-QuestionSP dataset.

As shown in Table~\ref{tab:prior-results}, the \system agent achieves new SOTA on the three QALD datasets and comes within 1.6\% F1 of the fine-tuned SOTA (WikiSP) on WikiWebQuestion. In particular, our 0-shot agent outperforms WikiSP by 24.2\% EM and 31.0\% F1 on Qald-7 (Task 4), showing its impressive cross-dataset generalization capability compared to a fine-tuned model. Our approach also outperforms the GPT-4-based ToG~\cite{sun2024thinkongraph} by 10.0\% EM on the same subset of QALD-10 used for evaluation in its paper.

\subsection{Results on the \dataset Dataset}
\label{sec:prior-works-on-spinach}

We benchmark the performance of four baseline systems on the \system dataset. These baselines are: (1) directly asking GPT-4o to answer the question without writing SPARQL, (2) directly asking GPT-4o to write a SPARQL query, (3) WikiSP~\cite{xu-etal-2023-fine} which is a fine-tuned LLaMA (7B-parameter) model, and (4) the GPT-4-based ToG agent~\citep{sun2024thinkongraph}. 

WikiSP expects a Named Entity Disambiguation (NED) module to predict the relevant entities given a question. The original NED module was based on a fine-tuned version of ReFinED~\citep{ayoola-etal-2022-refined}. To make the task easier for WikiSP, we directly use the gold entities that appear in the annotated SPARQL as inputs. Similarly, the ToG~\citep{sun2024thinkongraph} pipeline uses LLM to extract ``topic entities'' from the input question as the starting point of its knowledge graph exploration. We observed that using all the gold entities as topic entities, the model was unable to return any exploration paths for the first 80 questions in the validation set and simply defaulted to GPT-4. To make the task even easier for ToG, we sample up to 20 entities from the results of the gold SPARQL queries as the topic entities. Additional details on our setup and comparison of baseline systems can be found in Appendix~\ref{appendix:eval-baseline}.

As shown in Table~\ref{tab:prior-works-on-spinach}, the \system agent drastically outperforms all baselines. 
In particular, we highlight that the low accuracy of the GPT-4o QA system shows that this dataset mainly consists of long-tail knowledge unfamiliar to LLMs. For reference, GPT-4 was able to achieve 90.5\% EM on WebQuestionSP~\citep{tan2023chatgpt}. Although \system was able to achieve the SOTA on this dataset, the relatively lower performance compared to those achieved on previous dataset indicates that there is a lot of room for improving KBQA systems in the future.


\subsection{Error Analysis}

We randomly sample 20 cases on the \system dataset where the \system agent achieves less than 0.05 $F_1$ and conduct an error analysis. We observe that:
\begin{itemize}[topsep=0.5em]\itemsep-0.5em 
    \item \textbf{Property-related problems}: 40\% of errors are due to problems with properties. This includes cases when the \system agent fails to fetch the correct property or incorrectly uses a property (e.g. using it as a \texttt{wdt:} relation as opposed to the correct \texttt{ps:} or \texttt{pq:} qualifier).
    \item \textbf{Complicated SPARQL}: 30\% of errors are due to the failure of the \system agent to write complex SPARQL to fetch results, e.g., a complicated filter for finding people born in a specific month.\footnote{An interesting observation is that in 2 cases, GPT-4o ended the reasoning with ``I will fetch all related records and proceed to filter the results in Python'' after observing its previous SPARQL returned no results, suggesting there may be benefits to leveraging programming languages that are more familar to LLMs.}
    \item \textbf{Not enough exploration}: 15\% of errors are due to insufficient exploration performed by the LLM after reaching the maximum allowed number of actions. 
    \item \textbf{Inaccurate semantic parsing}: 10\% of errors are due to the LLM injecting an extra clause. For instance, when asked ``what items were published in ...'', the LLM assumes that the results have to be an instance of the domain entity ``book'', but there are many other types of results returned by the gold query (e.g. ``printed matter'' and ``legal act'').
    \item \textbf{Formatting issues}: 5\% of errors are due to format errors. An example is the LLM returning the date in full (``June 23 2021'') when only the  year is requested (``2021'').
\end{itemize}

\subsection{Ablation Study and Analysis}

We created the set of agent actions from our own experience in writing SPARQL.  
To understand how each action contributes to the performance of the \system agent, we remove the actions one at a time (except \textbf{\texttt{execute\_sparql}} and \textbf{\texttt{stop}}, without which the agent would never return a SPARQL query). The results on the \dataset dev set are reported in Table \ref{tab:ablation-study}. This ablation shows that all actions together contribute to the performance of the \dataset agent. We also report the distribution of the number of actions the \dataset agent took to answer questions from the \dataset dev set in Table \ref{tab:number-of-turns-spinach-agent} and the distribution of the number of tokens in Appendix \ref{sec:token-distribution}.

\begin{table}[!ht]
\small
    \centering
    \begin{tabular}{l|rr}
        \toprule
    & EM & F1 \\
    \midrule
\dataset agent &\textbf{21.4} &\textbf{46.4} \\
w/o \textbf{\texttt{get\_wikidata\_entry}} &11.7 &36.4 \\
w/o \textbf{\texttt{get\_property\_examples}} &10.4 &29.4  \\
w/o \textbf{\texttt{search\_wikidata}} &4.6 &25.3  \\
        \bottomrule
    \end{tabular}
    \caption{Ablation study of \dataset agent without each of its actions on the \dataset dev set.} 
\label{tab:ablation-study}
\end{table}

\begin{table}[!ht]
\small
    \centering
    \begin{tabular}{rr}
        \toprule
\# of actions & Percentage \\
    \midrule
3-5 &	28.6\% \\
6-8 &	33.8\% \\
9-11 &	11.0\% \\
12-14 &	7.1\% \\
15 &	19.5\% \\
        \bottomrule
    \end{tabular}
    \caption{Distribution of the number of actions (without counting rollbacked actions) the \dataset agent takes to answer a question from the \dataset dev set.} 
\label{tab:number-of-turns-spinach-agent}
\end{table}

\section{Conclusion}
 We propose the expert-annotated \dataset dataset drawn from real-world queries and introduce the \system agent that mimics how a human expert writes SPARQL queries. Experiments on prior works show that the \system agent achieves the new SOTA on 3 QALD datasets and comes within 1.6\% F1 to the fine-tuned SOTA model on WikiWebQuestions.
On the new \system dataset, our agent outperforms the best LLM agent approach, ToG, by 35.68\% $F_1$ on the test set.
The performance of 16.4\% EM and 45.3\% $F_1$ on the test set suggests ample opportunity for further improvement. Fortunately, as our \system agent diaplays all its intermediate steps, users can continue the conversation and revise their queries to help the agent derive the answer. We have deployed the \system agent at \url{https://spinach.genie.stanford.edu/} and on Wikidata at \url{https://www.wikidata.org/wiki/User:SpinachBot} as a community resource to facilitate the access of data in Wikidata. 



\section*{Limitations}
Since \system agent makes multiple LLM calls for each question, its latency and cost are higher compared to simpler systems. Other multi-stage KBQA pipelines with LLMs like \citet{sun2024thinkongraph} suffer from the same limitations. This seems to be the price for a more accurate KBQA system. However, recently, model distillation has been shown to be effective in improving model efficiency~\cite{semnani-etal-2023-wikichat}. We leave the exploration of this direction for future work.

As observed in recent works~\citep{kamalloo-etal-2023-evaluating, zhang2024spaghetti}, automatic metrics such as EM and $F_1$ often fail to accurately measure the capabilities of LLM-based systems. By verbalizing the SPARQL queries as accurately as possible, we aim to minimize the difference between automatic metrics and manual inspection. We thus expect these automatic metrics to still play a pivotal role in the evaluation of future work on the \dataset dataset.

Annotated datasets can also suffer from annotation errors and mistakes~\citep{zang-etal-2020-multiwoz}. We address this limitation by having experts annotate the dataset and independently cross-validating each other's annotations, as discussed in Appendix~\ref{appendix:annotation-details}.

\section*{Ethical Considerations}
We release the \dataset dataset in this paper. This dataset is built using publicly available data in the Wikidata Request Query forum, which is licensed under the Creative Commons CC0 License. We don't expect any harm being produced from the usage of this dataset. The generation and annotation of the dataset were done by the authors of this paper.

No GPU-based training was carried out in our experiments, as we mainly used the OpenAI API to call the GPT-4o model. To evaluate the LLaMA baseline from \citet{xu-etal-2023-fine}, we used a Linux server with one NVIDIA A100 GPU for <1 hour.

To facilitate further research, we release our code and data for the \system agent and the \dataset dataset. The \system agent is released under Apache License, version 2.0. The \dataset dataset, derived from the Wikidata Request a Query forum, is released under the CC BY-SA 4.0 license, the same license that covers the forum.

\section*{Acknowledgements}

We thank Isaac Johnson at the Wikimedia Foundation for helpful discussions about this work, Harshit Joshi for his help with the project, the ACL ARR reviewers for their valuable comments and suggestions, and members of the Stanford Open Virtual Assistant Lab (OVAL) and NLP group for their feedback and support. This work is supported in part by the Verdant Foundation, Microsoft Azure AI credits, KDDI, and the Stanford Human-Centered Artificial Intelligence (HAI) Institute.

\bibliography{anthology, custom}

\appendix

\section{Definition of metrics used in Table \ref{tab:dataset-analysis}}
\label{appendix:dataset-metrics-definition}

For each query, we define the number of clauses as the number of atomic nodes of a SPARQL abstract syntax tree (AST), where an atomic node is defined as one of: a projection clause (i.e., \texttt{SELECT ... WHERE}), a single subject-relation-object clause, a group by clause, a having by clause, a filter clause, a sorting clause, a \texttt{MINUS} clause, and a join clause. The number of projections is defined as the number of fields in the final \texttt{SELECT ... WHERE} clause. The number of relations is defined as the number of subject-relation-object clauses. The number of subjects is defined to be the number of uniquely occuring subjects (either variables or entities) in each subject-relation-object clause. The number of predicates is defined to be the unique number of properties (i.e. a PID of the form \texttt{P...}). The number of objects is defined to be the number of uniquely occuring objects (either variables or entities) in each subject-relation-object clause, where we also count the number of \texttt{y} occuring in each \texttt{VALUES} clause of the form \texttt{VALUES ?x \{y\}}. The number of literals is defined to be the unique number of strings (e.g. \texttt{``wikidata''}) or numerical numbers (e.g. \texttt{123.45}). The number of unique properties is the number of Wikidata properties across all queries in a dataset (e.g. \texttt{P123}). The number of unique forms is the number of unique query patterns that occur in a dataset when one ignores all unique query variables, property IDs (PIDs), entity IDs (QIDs), numbers, and string literals and simply counts the resulting patterns.

Similar to prior work~\citep{grailQA}, we use hand-crafted rules (including regular expressions) to retrieve these metrics.

\section{Additional Comparison between \dataset and Prior KBQA Datasets}
\label{appendix:additional-comparison-with-synthetic-data}

Datasets with synthetically generated SPARQL queries typically attempt to cover the space of possible SPARQLs using heuristics, resulting in relatively easy SPARQL structures repeated over and over again. To quantitatively show this difference, we calculate the number of unique query patterns similar to what was proposed for SQL in \citet{finegan-dollak-etal-2018-improving}. Specifically, we de-duplicate SPARQL queries after ignoring all query variables, property IDs (PIDs), entity IDs (QIDs), numbers, and string literals and count the resulting patterns. Table \ref{tab:additional-comparison-with-synthetic-data} shows the comparison of the number of unique query patterns for \dataset and several other KBQA datasets. As shown here, even though synthetic datasets such as GrailQA, KQA-Pro, and CWQ contain many more data points, a small portion of query patterns are repeated over and over again. \dataset, on the other hand, features organic queries found in the wild where each query is structurally unique and poses a new challenge.

\begin{table}[ht]
\small
\centering
\begin{tabular}{lrrr}
\toprule
Dataset & Size & UQPs & UQPs  \\
       &              &            & / Size  \\
\midrule
GrailQA (train+dev)    & 51100   & 116  & 0.227\% \\
KQA-Pro                & 117970  & 1689 & 1.432\% \\
CWQ                    & 34689   & 402  & 1.159\% \\
WikiWebQuestions        & 4316    & 176  & 0.041\% \\
\dataset                & 320     & 320  & 100.0\% \\
\bottomrule
\end{tabular}
\caption{Comparison of Unique Query Patterns (UQPs) in \dataset and prior works.}
\label{tab:additional-comparison-with-synthetic-data}
\end{table}

We also report the number of unique properties in SPARQL queries of the \dataset dataset and prior Wikidata datasets in Table \ref{tab:additional-comparison-properties}. Datasets that repeat the same few properties over and over again have a low diversity of logical forms. Datasets that synthesize their logical forms like MCWQ especially have a low count of unique properties and do not properly test the ability of systems to explore the knowledge graph as they can simply memorize all of them during training. As shown here, SPINACH contains the most unique properties, and is quite diverse in this sense.

\begin{table}[ht]
\small
\centering
\begin{tabular}{lrr}
\toprule
Dataset            & \# UPs & Dataset Size \\
\midrule
MCWQ               & 27  & 124187 \\
QALD 9             & 158 & 507 \\
QALD 10            & 177 & 394 \\
RUBQ               & 251 & 2910 \\
WikiWebQuestions    & 189 & 4316 \\
\dataset            & 298 & 323 \\
\bottomrule
\end{tabular}
\caption{Comparison of number of Unique Properties (UPs) in \dataset and prior works.}
\label{tab:additional-comparison-properties}
\end{table}

\section{Additional Details on \dataset Dataset}
\label{appendix:additional-detail-dataset-collection}

\subsection{Additional Details on the Request a Query Forum}
\label{appendix:additional-detail-request-a-query}

We have found that posters on the Request a Query forum generally come to the forum with a clear understanding of what they would like to query, or even have a sample SPARQL query in a different field that they would like to adapt (e.g. converting a query about basketball statistics to baseball statistics, or converting a query about museums in the San Francisco area to operas in the Paris). In other words, most questions are already very complex in the first post on the forum.

We have analyzed the source conversations of data points in the \dataset dataset to derive the distribution of conversation lengths, where longer conversations may have more back-and-forth between users who are asking questions and users who are attempting to respond to those questions. We exclude conversations from the Request a Query forum that have no responses, as well as conversations that do not contain any SPARQL queries as there is no good candidate SPARQL to use for annotation. Shorter conversations (1-3 responses) comprise 57\% of the dataset, medium length conversations (4-6 responses) comprise 31.9\% of the dataset, and longer conversations (7 or more responses) comprise 11.1\% of the dataset. A detailed breakdown is available in Table \ref{tab:detail-breakdown-forum-conversations}. The length of a conversation is defined to be the number of responses posted.

\begin{table}[!ht]
\small
    \centering
    \begin{tabular}{rr}
        \toprule
    length of conv. & percentage \\
    \midrule
1      & 13.9\% \\
2      & 14.9\% \\
3      & 28.2\% \\
4      & 13.3\% \\
5      & 13.0\% \\
6      & 5.6\% \\
7      & 4.0\% \\
8      & 1.9\% \\
9      & 2.2\% \\
>10    & 3.1\% \\
        \bottomrule
    \end{tabular}
    \caption{Distribution of the length of conversations for each of data point in the \dataset.} 
\label{tab:detail-breakdown-forum-conversations}
\end{table}

\subsection{Discussion on Size and Statistical Power of \dataset}
\label{appendix:size-discussion}

NLP datasets are often used to differentiate between two proposed systems, for example, to determine if a new system outperforms the state-of-the-art or not. The size of the validation/test set determines the minimum detectable effect (MDE)~\citep{card-etal-2020-little}, i.e. the minimum improvement in a metric (like exact match) that will yield sufficiently powered comparisons. Following the approach in \citet{card-etal-2020-little}, we estimate that the SPINACH dev set is sufficiently large to differentiate between the \dataset agent and systems that are 9.0\% or more better than it. Meaning, that if a future paper presents a system scoring 30.4\% EM or higher on the dev set of \dataset, their experiment will be statistically significant with the commonly used significance level (alpha) of 5\%. Separately, we also note that all the improvements we report compared to our baselines in Table \ref{tab:prior-works-on-spinach} are statistically significant.

Many influential datasets in the LLM era contain fewer examples than traditional ML datasets, often because they are only meant to serve as a validation and test set instead of a training set. The \dataset dataset calls for systems that can dynamically explore large and often incomplete KB schemas and reason about them, as opposed to relying on training data. The poor performance of \citet{xu-etal-2023-fine}, a model fine-tuned on a larger training set, showcases that models fine-tuned on often larger datasets do not necessarily generalize well (Table \ref{tab:prior-works-on-spinach}). Furthermore, many recent LLM-based systems (e.g. one of our baselines, ToG~\citep{sun2024thinkongraph}) only evaluate on smaller datasets, or small subsets of larger datasets, due to the high cost of LLM APIs.

For these reasons, we believe the SPINACH dataset is a good middle ground that keeps the cost of our expert annotation and running multiple experiments in future papers manageably low, while being reasonably powered to differentiate between and track the progress of future systems that use this dataset.

\subsection{Additional Details on Annotation}

\label{appendix:annotation-details}
\textbf{Expert annotations}: The \dataset dataset is annotated by 3 authors of this paper who are experts with extensive knowledge in SPARQLs and Wikidata. The 3 experts first engaged in a long discussion with specific examples to standardize the annotation procedure. 2 experts first annotated the dev and test sets, with the 3rd expert double-checking and validating the annotations.

\textbf{LLM suggestions}: To facilitate the annotation process, we used GPT-4o to generate preliminary annotation suggestions to the experts. For each example, the experts are shown with (1) the original SPARQL query and with properties \& entities substituted with labels, (2) link to the specific forum discussion, (3) two LLM-suggested verbalizations (one more verbose and one more natural), and (4) an LLM suggestion of whether to modify, include, or exclude the query with its reasonings. For each data example, the experts executed numerous intermediate SPARQL queries on the Wikidata site to verify the query's validity and finalize the annotated SPARQL.

\textbf{Automatic exclusion of mwapi in queries}: Some SPARQL queries on the forum make use of \texttt{mwapi}\footnote{\url{https://en.wikibooks.org/wiki/SPARQL/SERVICE_-_mwapi}}. The usage of these APIs are mostly for optimization or are otherwise very specific to the structure of Wikimedia sites. We note that sometimes, removing these APIs would lead to small changes in the results (most likely due to the outdated discrepancy between the API outputs and Wikidata-direct outputs). To standardize the dataset and avoid issues during evaluation, we use regex to delete all occurrences of \texttt{mwapi}s in the SPARQLs before executing them.

\subsection{Examples of queries modified}
\label{appendix:examples-of-queries-modified}
The following are examples of how we modify target SPARQLs.

\textit{Wikimedia presentation queries}: \href{https://www.wikidata.org/wiki/Wikidata:Request_a_query/Archive/2020/07#Two-letter_genera,_on_Wikispecies}{This dicsussion}, titled ``Two-letter genera, on Wikispecies'' in July of 2020, involves the following snippets of retrieving information from \texttt{species.wikimedia.org} for certain taxa:

\begin{lstlisting}[basicstyle=\ttfamily\small]
  ?wikispecies schema:about ?item .
  ?wikispecies schema:isPartOf <https://species.wikimedia.org/> .
\end{lstlisting}

This, however, is very specific to the structure of different Wikimedia sites. Removing these two clauses does not modify the core parts of the meaning. We thus remove these two clauses and the corresponding target \texttt{?wikispecies} in the projection. The resulting SPARQL is then included in the \system validation set.

\textit{Queries obscured by optimizations}: \href{https://www.wikidata.org/wiki/Wikidata:Request_a_query/Archive/2020/09#Persons_from_a_certain_time_period_and_country}{This discussion}, titled ``Persons from a certain time period and country'' in Septempter of 2020, contains a query that makes use of the following optimization:

\begin{lstlisting}[basicstyle=\ttfamily\small]
  int:Prior hint:rangeSafe true .
\end{lstlisting}

which is used to speed up the succeeding filter. Removing this optimization in this case does not significantly increase SPARQL executing time. We thus remove this clause and include the example in the \system validation set.

\textit{Formatting clauses}:  
\href{https://www.wikidata.org/wiki/Wikidata:Request_a_query/Archive/2020/08#Query_all_taxons_which_are_trees}{This discussion}, titled ``Query all taxons which are trees'', results in a SPARQL of the following structure:

\begin{lstlisting}[basicstyle=\ttfamily\small]
SELECT
  ?taxon ?sample (GROUP_CONCAT(DISTINCT str(?commonname); separator = "//") as ?commonnames) 
WHERE
{
...
}
GROUP BY ?taxon ?sample
\end{lstlisting}

Using a group by, it is trying to concatenate all \texttt{commonnames} into one single string, separated by \texttt{"//"}. However, it is very difficult to accurately capture this in natural language, and a system could return the results in different orders, raising issues for evaluation. We thus exclude this concatenation and instead change it to a counting operation:

\begin{lstlisting}[basicstyle=\ttfamily\small]
SELECT
  ?taxon ?sample (COUNT(?commonname) as ?commonnamecount)
WHERE
{
...
}
GROUP BY ?taxon ?sample
\end{lstlisting}
The resulting SPARQL is then included in the \system validation set.






\begin{table*}[ht]
\small
    \centering
    \begin{tabular}{llp{10.5cm}}
        \toprule
Percentile   & Page Views Per Year & Example question in the \dataset dev set \\
    \midrule
10th & 3,499,391 & Who are citizens of \textbf{Norway} that have held a position as a teacher or a subclass of teacher, along with their labels, dates of birth (if available), and Norwegian historical register of persons IDs (if available)? \\ \midrule
25th & 873,067 & What are the \textbf{Nazi concentration camps} or subcamps? Include the following way of finding such a camp: (1) those that are classified as a \textbf{Nazi concentration camp}, (2) those that are considered the \textbf{subcamps}, and (3) \textbf{subsidiaries of Nazi concentration camps}. For each one, also find its coordinates (e.g. literals such as Point(9.1978 49.1686)). Each pair of camp and coordinate should only appear once. \\ \midrule
50th & 131,131 & What are the religious buildings located in the region of \textbf{Molise, Italy}? \\ \midrule
75th & 	44 & 	What are the current municipal districts, urban districts in Russia, or \textbf{administrative districts of Moscow} located within \textbf{Perm Krai}? \\ \midrule
90th & 0 & 	Who are the members of the historic \textbf{Lower House} and Upper House \textbf{of the Parliament of Sweden}, whose party's Swedish label contains the string "vilde" in their party name? For each person, also return the associated party, and the start (if available) and end (if available) times of the party. \\
        \bottomrule
    \end{tabular}
    \caption{Distribution of page views for the entity's English Wikipedia page in July 2023 - June 2024 for each entity in \dataset's gold SPARQL queries.} 
\label{tab:analysis-popularity-of-queries}
\end{table*}

\subsection{Examples of queries excluded}
\label{appendix:examples-of-queries-excluded}
The following are examples of when, after modifications, no meaningful part of the SPARQL remains. The discussions are thus excluded from our dataset.

\textit{Wikimedia presentation queries}: \href{https://www.wikidata.org/wiki/Wikidata:Request_a_query/Archive/2020/04#?item_=wd:QXXX}{The discussion}, titled ``?item =wd:QXXX'' in April of 2020, results in a SPARQL that is only trying to fetch wikipedia pages that are about the item ``Gambling, Gods And LSD''. The usage of \texttt{schema:about} property is very specific to Wikimedia and could raise confusion on the meaning of ``about''. It is thus excluded. \href{https://www.wikidata.org/wiki/Wikidata:Request_a_query/Archive/2018/03#Bengali_Wikipedia_articles_with_no_Wikidata_statement}{The discussion}, titled ``Bengali Wikipedia articles with no Wikidata statement
'' in March of 2018, results in a SPARQL trying to fetch Wikidata items with no statements that are the topic of discussion of Bengali Wikipedia articles. The usage of \texttt{wikibase:statements}, \texttt{schema:about}, \texttt{schema:isPartOf}, and \texttt{wikibase:sitelinks} properties are all very specific to the structure of Wikimedia sites and can create confusions for KBQA systems. It is thus excluded.

\textit{Questions on complex SPARQL code}: 
\href{https://www.wikidata.org/wiki/Wikidata:Request_a_query/Archive/2021/03#Scatterplot_query}{This discussion}, titled ``Scatterplot query'' in March of 2021, is an example where the original requester comes in with a SPARQL that is complicated whose meaning is already difficult to acurately express in English. The final SPARQL from the conversation (hyperlinked in `` a bit more baroque, with axes'') only adds to the complexity.

\textit{Queries obscured by optimizations}:
\href{https://www.wikidata.org/wiki/Wikidata:Request_a_query/Archive/2019/06#P31/wdt:P279*_wd:Q16917_in_wdt:P131/wdt:P131*_wd:Q25_(query_optimization)}{This discussion}, titled ``Section ``P31\/wdt:P279\* wd:Q16917 in wdt:P131\/wdt:P131\* wd:Q25 (query optimization)'', involves only the refactoring of a query into using two sub-queries which are then joined together to avoid timing outs. The two queries are semantically equivalent. Due to reproducibility challenges from this refactoring, this discussion is excluded.

\textit{Formatting clauses}:  
\href{https://www.wikidata.org/wiki/Wikidata:Request_a_query/Archive/2022/02#preferred_format_for_id}{The discussion}, titled ``preferred format for id'' in Feburary of 2022, only contains discussions on differnt ways for string processing in SPARQL and is thus excluded. \href{https://www.wikidata.org/wiki/Wikidata:Request_a_query/Archive/2018/06#Custom_link_formatting_in_WDQS_results}{The discussion}, titled ``Custom link formatting in WDQS results'' in June of 2018, only contains results in a SPARQL that focuses on converting string formats after minimal use of Wikidata (fetching the sitelinks of one item) and is thus excluded.

\subsection{Analysis of the Popularity of Queries}
\label{appendix:analysis-popularity-of-queries}

Following the method in \citet{semnani-etal-2023-wikichat} and \citet{mallen-etal-2023-trust}, we estimate and report the popularity of queries in the \dataset dataset. Specifically, for each unique entity in the gold SPARQL queries of the \dataset dataset, we obtain the number of page views for its English Wikipedia page in the past year (July 2023 - June 2024), counting those without an English Wikipedia page as 0. We show in Table \ref{tab:analysis-popularity-of-queries} a distribution of the page views, sorted from higher to lower page views, where the gold entity is shown in bold. As shown here, the dataset contains a wide range of entities, from popular ones like \textbf{Norway} at 3,606,300 page views per year to tail entities like \textbf{Administrative divisions of Moscow} at 44 page views per year.

\section{System and Evaluation Details}
\label{appendix:eval-baseline}
Unless otherwise specified, LLMs are used with greedy decoding, i.e. with $temperature=0$, with the exception of \system agent's policy prompt, which is run with $temperature=1$ and nucleus sampling~\cite{holtzman2019curious} with $p=0.9$

The ToG agent by default builds a local version of Wikidata using the \texttt{simple-wikidata-db} library\footnote{\href{https://github.com/neelguha/simple-wikidata-db}{Github: neelguha/simple-wikidata-db}}. However, building this index from scratch is extremely computationally expensive. \citet{sun2024oda} report that they need to deploy the Wikidata dump across five AWS EC2 instances, each consisting of a 768GB machine with 48 cores. \citet{Fahl_Holzheim_Westerinen_Lange_Decker_2022} reported various other methods of hosting Wikidata locally. We attempted using Qlever as the SPARQL Engine but failed to do so on the latest Wikidata dump. For the evaluation of ToG, we re-implemented the same logic using dynamic Wikidata API calls.

We adopted the same hyperparameters for ToG as used in the original paper. For the experiment in Section \ref{sec:prior-works-on-spinach}, we used D=3 and T=3 as the hyperparameters for graph exploration.

The GPT-4 and ToG systems return results in strings of the entities instead of entity IDs; we convert the gold SPARQL output to only contain their English labels and calculate EM and F1 metrics.

\subsection{Token Distribution for the \dataset Agent}
\label{sec:token-distribution}

\begin{table}[!ht]
\small
    \centering
    \begin{tabular}{rr}
        \toprule
\# of Tokens & Percentage \\
    \midrule
4912.50 - 10330.50 & 16.23\% \\
10330.50 - 15748.50 & 24.03\% \\
15748.50 - 21166.50 & 18.18\% \\
21166.50 - 26584.50& 8.44\% \\
26584.50 - 32002.50 & 6.49\% \\
32002.50 - 37420.50 & 5.84\% \\
37420.50 - 42838.50 & 10.39\% \\
42838.50 - 48256.50& 7.14\% \\
48256.50 - 53674.50 & 1.30\% \\
53674.50 - 59092.50 & 1.95\% \\
        \bottomrule
    \end{tabular}
    \caption{Distribution of total input tokens needed to answer a question (including all intermediate steps) from the \dataset dev set.} 
\label{tab:input-token-distribution}
\end{table}

\begin{table}[!ht]
\small
    \centering
    \begin{tabular}{rr}
        \toprule
\# of Tokens & Percentage \\
    \midrule
164.00 - 445.30 & 33.77\% \\
445.30 - 726.60 & 27.92\% \\
726.60 - 1007.90 & 14.29\% \\
1007.90 - 1289.20& 11.04\% \\
1289.20 - 1570.50 & 6.49\% \\
1570.50 - 1851.80 & 3.90\% \\
1851.80 - 2133.10 & 1.30\% \\
2133.10 - 2414.40& 0.65\% \\
2414.40 - 2695.70 & 0.00\% \\
2695.70 - 2977.00 & 0.65\% \\
        \bottomrule
    \end{tabular}
    \caption{Distribution of total output tokens needed to answer a question (including all intermediate steps) from the \dataset dev set.} 
\label{tab:output-token-distribution}
\end{table}

Table \ref{tab:input-token-distribution} and Table \ref{tab:output-token-distribution} show the distribution of input and output tokens needed to answer questions from the \dataset dev set, respectively. On average, running SPINACH agent on each question costs \$0.1253 using GPT-4o.

\section{Prompts used in the \system system}

 The policy prompt and the prompt used to prune the output of \textbf{\texttt{get\_wikidata\_entry()}} are shown in Table \ref{prompt:controller} and Table \ref{prompt:prune}, repsectively.

\begin{table*}
\begin{lstlisting}[basicstyle=\ttfamily\tiny]
# instruction
Your task is to write a Wikidata SPARQL query to answer the given question. Follow a step-by-step process:

1. Start by constructing very simple fragments of the SPARQL query.
2. Execute each fragment to verify its correctness. Adjust as needed based on your the observations.
3. Confirm all your assumptions about the structure of Wikidata before proceeding.
4. Gradually build the complete SPARQL query by adding one piece at a time.
5. Do NOT repeat the same action, as the results will be the same.
6. The question is guaranteed to have an answer in Wikidata, so continue until you find it.
7. If the user is asking a True/False question with only one answer, use ASK WHERE to fetch a True/False answer at the very end.
8. In the final SPARQL projections, do not only ask for labels. Ask for the actual entities whenever needed (e.g. instead of doing `SELECT xLabel`, do `SELECT x`).
9. If the final result was contained in last round's `get_wikidata_entry` and you are ready to stop, use `execute_sparql` and generate a SPARQL to retrieve that results.

Form exactly one "Thought" and perform exactly one "Action", then wait for the "Observation".

Possible actions are:

- get_wikidata_entry(QID): Retrieves all outgoing edges (linked entities, properties, and qualifiers) of a specified Wikidata entity using its QID.
- search_wikidata(string): Searches Wikidata for entities or properties matching the given string.
- get_property_examples(PID): Provides a few examples demonstrating the use of the specified property (PID) in Wikidata.
- execute_sparql(SPARQL): Runs a SPARQL query on Wikidata and returns a truncated result set for brevity.
- stop(): Marks the last executed SPARQL query as the final answer and ends the process.


# input
Question: {{ question }}

{% if action_history %}
{% for i in range(0, action_history|length) %}

{{ action_history[i] }}
{% endfor %}
{% endif %}

Output one "Thought" and one "Action":


\end{lstlisting}
\caption{The policy prompt of the \system agent.}
\label{prompt:controller}
\end{table*}

\begin{table*}
\begin{lstlisting}[basicstyle=\ttfamily\tiny]
# instruction
At each turn, you are given a Wikidata entry and a question.
You want to write a SPARQL query that answers the question.
As the first step, remove the parts of the Wikidata entry that could not be potentially helpful when writing the SPARQL.
The output should be a json object containing part of the Wikidata entry.

# few-shot example 1, input
Wikidata entry for OneRepublic (Q1438730, 'OneRepublic' is an American pop rock band formed in Colorado Springs, Colorado, in 2002):
{
  "instance of (P31)": "musical group (Q215380)",
  ...
  "social media followers (P8687)": {
    "3134158": {
      "Qualifiers": [
        {
          "point in time (P585)": "4 February 2023"
        }
      ]
    },
    "3276596": {
      "Qualifiers": [
        {
          "point in time (P585)": "6 January 2021"
        }
      ]
    },
    "3178896": {
      "Qualifiers": [
        {
          "point in time (P585)": "2 March 2022"
        }
      ]
    },
    "3720919": {
      "Qualifiers": [
        {
          "point in time (P585)": "10 May 2018"
        }
      ]
    }
  }
}

Question: "What is the title of the second single on OneRepublic's third album Native?"

# few-shot example 1, output
{
  "instance of (P31)": "musical group (Q215380)",
  "record label (P264)": "Columbia Records (Q183387)",
  "discography (P358)": "OneRepublic discography (Q935670)",
}



# few-shot example 2, input
Wikidata entry for Barack Obama Sr. (Q649593, Barack Obama Sr. was an economist and the father of Barack Obama jr.):
{
  "place of birth (P19)": {
    "Nyang'oma Kogelo (Q1027168)": {
      "Qualifiers": [
        {
          "country (P17)": "Kenya Colony"
        }
      ]
    }
  },
  ...
}

Question: "where did barack obama sr. attend school?"

# few-shot example 2, output
{
  "place of birth (P19)": {
    "Nyang'oma Kogelo (Q1027168)": {
      "Qualifiers": [
        {
          "country (P17)": "Kenya Colony"
        }
      ]
    }
  },
  "place of death (P20)": "Nairobi (Q3870)",
 ...
}


# input
Wikidata entry for "{{ entity_and_description }}":
{{ outgoing_edges }}

Question: "{{ question }}"
\end{lstlisting}
\caption{The 2-shot prompt used to prune the output of \texttt{get\_wikidata\_entry() action}. Here we omit parts of the few-shot example inputs for space.}
\label{prompt:prune}
\end{table*}

\section{Full Example Trace of the \system Agent}

Listing \ref{fig:full_example} contains a full sequence of thoughts, actions and observations of the \system agent for the given input question.

\onecolumn
\begin{lstlisting}[language=ReAct, caption={Full version of the example in Figure~\ref{fig:example}}, label={fig:full_example}]
"""
Question: SteplanI hav
"""

Thought: First, I need to identify the properties and entities related to people affiliated with the University of Washington School of Music, their educational background, and the musical instruments they play. This will help in formulating the final SPARQL query.
>> search_wikidata("University of Washington School of Music")
Observation: University of Washington School of Music (Q98035717): school of music

Thought: Now that I have identified the entity for the University of Washington School of Music, I need to find the properties that can help me identify people affiliated with it and the properties related to their education and musical instruments.
>> get_wikidata_entry("Q98035717")
Observation: 
  {
    instance of (P31): {
      faculty (Q180958): {
        Qualifiers: [
          {
            of (P642): University of Washington (Q219563)
          }
        ]
      },
      music school (Q1021290): {}
    },
    field of work (P101): [
      higher education (Q136822),
      postgraduate education (Q141212),
      musicology (Q164204),
      music (Q638),
      research (Q42240),
      ethnomusicology (Q208365)
    ],
    located in the administrative territorial entity (P131): Seattle (Q5083),
    location (P276): Music Building (Q98690890),
    has subsidiary (P355): University of Washington Laboratory for Music Cognition, Culture & Learning (Q101157561),
    part of (P361): University of Washington College of Arts and Sciences (Q7896566),
    has part(s) (P527): University of Washington Laboratory for Music Cognition, Culture & Learning (Q101157561),
    parent organization (P749): University of Washington College of Arts and Sciences (Q7896566)
  }

Thought: To find people affiliated with the University of Washington School of Music, I need to look for properties that define such an affiliation. Additionally, I need to identify properties related to their education at the University of Washington and the musical instruments they play. I will start by searching for the property that links people to their affiliations.
>> search_wikidata("affiliation")
Observation: 
  - affiliation (Q46135267)
    connection, formally or informally, with an organization or intellectual movement
  - filiation (Q5861944)
  system for recognizing the lineage or kinship of a person in anthropology
  - affiliation (Q2620373)
    archaic term for the legal establishment of paternity
  - affiliated entity (Q69467106)
    inverse property label item for P1416
  - Affiliation (Q93774359)
    scholarly article
  - Affiliations Committee (Q46815761)
    Wikimedia community-run committee
  - Wikimedia Affiliations Committee wiki (Q33120941)
    Wikimedia private wiki
  - Affiliation and participation (Q107433952)
    master's thesis of Hannah Marie Mayfield
  - affiliation (P1416)
    organization that a person or organization is affiliated with (not necessarily member of or employed by)
    Data Type: wikibase-item
  - affiliation string (P6424)
    qualifier to provide the published string form of affiliation attached to an author
    Data Type: string
  - Olympedia affiliations ID (P9447)
    identifier for the affiliation of athletes in Olympedia athlete database
    Data Type: external-id (Represents an identifier used in an external system.)

Thought: The property `affiliation (P1416)' seems to be the most relevant for linking people to the University of Washington School of Music. Next, I need to identify the property for their education at the University of Washington and the property that links people to the musical instruments they play.
>> search_wikidata("educated at")
Observation:
  - educated at (P69)
    educational institution attended by subject
    Data Type: wikibase-item

Thought: I have identified the property `educated at (P69)' to link people to the University of Washington. Next, I need to find the property that links people to the musical instruments they play.
>> search_wikidata("musical instrument")
Observation: 
  - musical instrument (Q34379)
    device created or adapted to make musical sounds
  - heraldic musical instrument (Q56877088)
    category of heraldic charges
  - Musical instrument (Q102413357)
    Oil of canvas by Alla Grigoryan
  - Musical Instrument (Q102413736)
    painting by Musical instrument
  - Musical instrument (Q86150836)
    sculpture in the church Brother Klaus in the city of Bern, Switzerland
  - Musical Instrument (Q80002888)
    musical instrument by unknown artist (1929.322)
  - Musical instrument (Q123046258)
    US patent 11482197
  - musical instrument maker (Q1955150)
    manufacturer of musical instruments
  - instrument (P1303)
    musical instrument that a person plays or teaches or used in a music occupation
    Data Type: wikibase-item
  - MIMO instrument ID (P3763)
    identifier for a musical instrument, on the instrument keyword thesaurus from Musical Instruments Museums Online
    Data Type: external-id (Represents an identifier used in an external system.)

Thought: I have identified the relevant properties: `affiliation (P1416)' to link people to the University of Washington School of Music, `educated at (P69)' to link people to the University of Washington, and `musical instrument (P1303)' to identify the instruments they play. I will now construct a SPARQL query that combines these properties to get the required information.
>> execute_sparql("""
    SELECT ?instrument (COUNT(*) AS ?count) WHERE {
      ?person wdt:P69 wd:Q1063349;  
              wdt:P1416 wd:Q98035717;  
              wdt:P1303 ?instrument.  
    }
    GROUP BY ?instrument
    """
)
Observation:
  Query returned empty result.

Thought: It appears that the query returned an empty result. This could be due to several reasons: either there are no people matching all criteria, or the properties might not be used exactly as I assumed. To ensure accuracy, I will individually verify if there are people affiliated with the University of Washington School of Music who have also been educated at the University of Washington.
>> execute_sparql(
    """
    SELECT ?person WHERE {
      ?person wdt:P69 wd:Q1063349;
        wdt:P1416 wd:Q98035717.
    }
    LIMIT 10
    """
)
Observation: Query returned empty result.

Thought: The absence of results could indicate that the properties or the specific entities used in the query might not be correct or commonly utilized in this context. I will check if there are people who have been educated at the University of Washington (Q1063349) and then see if I can refine the affiliation aspect.
>> execute_sparql(
    """
    SELECT ?person WHERE { ?person wdt:P69 wd:Q1063349. }
    LIMIT 10
    """
)
Observation: Query returned empty result.

Thought: Since my previous queries returned empty results, I need to verify if there are any individuals in Wikidata who have been educated at the University of Washington (Q1063349). This will help me confirm if I am using the correct properties and entities.
>> get_wikidata_entry("Q1063349")
Observation: 
  {
    country (P17): Russia (Q159),
    instance of (P31): hamlet (Q5084),
    located in the administrative territorial entity (P131): Tavtimanovsky selsoviet (Q4449505),
    population (P1082): {
      11: {
        Qualifiers: [
          {
            point in time (P585): 1 January 2010,
            determination method (P459): census (Q39825)
          },
          {
            determination method (P459): census (Q39825),
            point in time (P585): 1 January 2002
          }
        ]
      },
      9: {
        Qualifiers: [
          {
            point in time (P585): 1 January 2009,
            determination method (P459): census (Q39825)
          }
        ]
      }
    }
  }

Thought: The query results indicate that there may be an issue with the QID I used for the University of Washington (Q1063349). It seems to correspond to a location in Russia rather than the university. I should search for the correct QID for the University of Washington.
>> search_wikidata("University of Washington")
Observation: 
  - University of Washington (Q219563)
    public research university in Seattle, Washington, United States
  - University of Washington Collection (NAID 931) (Q59502962)
    collection in the National Archives and Records Administration's holdings
  - University of Washington Department of Chemistry (Q97958839)
    academic department at the University of Washington
  - University of Washington Department of Architecture (Q98844905)
    academic department at the University of Washington
  - University of Washington Department of Civil and Environmental Engineering (Q97990078)
    academic department at the University of Washington
  - University of Washington Department of Chemical Engineering (Q99196105)
    academic department at the University of Washington
  - University of Washington Paul G. Allen School of Computer Science & Engineering (Q98186807)
    school of the University of Washington
  - University of Washington School of Music (Q98035717)
    school of music

Thought: The correct QID for the University of Washington is Q219563. I should now update my SPARQL query to use this correct QID for identifying people who have been educated at the University of Washington and who are affiliated with the University of Washington School of Music.
>> execute_sparql(
    """
    SELECT ?person WHERE {
      ?person wdt:P69 wd:Q219563;
        wdt:P1416 wd:Q98035717.
    }
    LIMIT 10
    """
)
Observation: 
  | Person       |
  |--------------|
  | Q107692821   |
  | Q107705263   |
  | Q107706107   |
  | Q107707997   |
  | Q107708199   |
  | Q107708456   |
  | Q107720948   |
  | Q107773759   |
  | Q107782800   |
  | Q107789859   |

Thought: Now that I have identified individuals who have been educated at the University of Washington (Q219563) and are affiliated with the University of Washington School of Music (Q98035717), I will update my SPARQL query to find the musical instruments played by these individuals and count the occurrences of each instrument.
>> execute_sparql(
    """
    SELECT ?instrument ?instrumentLabel (COUNT(?student) AS ?count) WHERE {
      ?student wdt:P1303 ?instrument;
        wdt:P1416 wd:Q98035717;
        wdt:P69 wd:Q219563.
      SERVICE wikibase:label { bd:serviceParam wikibase:language "en". }
    }
    GROUP BY ?instrument ?instrumentLabel
    """
)
Observation: 
  | instrument   | instrumentLabel | count |
  |--------------|-----------------|-------|
  | Q5994        | piano           | 99    |
  | Q1467960     | mbira           | 2     |
  | Q8350        | trombone        | 11    |
  | Q8338        | trumpet         | 8     |
  | Q17172850    | voice           | 32    |
  | ...          | ...             | ...   |
  | Q302497      | mandolin        | 1     |
  | Q187851      | recorder        | 1     |
  | Q185041      | cor anglais     | 1     |
  | Q83509       | piccolo         | 1     |

Thought: The query successfully returned the musical instruments played by individuals who are affiliated with the University of Washington School of Music and have been educated at the University of Washington, along with the counts of each instrument. I will mark this query as complete.
>> stop()
\end{lstlisting}

\end{document}